\documentclass[journal]{IEEEtran}
\usepackage{enumitem}
\usepackage{multirow}
\usepackage[table,xcdraw]{xcolor}
\usepackage{graphicx}
\usepackage{amsmath}
\usepackage{amsfonts}
\usepackage{algorithm}
\usepackage{amsthm,amssymb}
\usepackage{mathrsfs}
\usepackage{algorithmic}
\usepackage{makecell}%解决公式在表格中位置太小的问题
\usepackage{easyReview}
\usepackage{booktabs}%解决表格横向粗细问题
\setcellgapes{3pt}

\setcellgapes{3pt}

\usepackage[justification=centering]{caption}
\usepackage{color}
\usepackage[subfigure]{tocloft}
\usepackage{subfigure}
\usepackage{tabularx}
\usepackage{amsmath, bm}
\usepackage{hyperref}
\hypersetup{hidelinks,
colorlinks=true,
allcolors=blue,
pdfstartview=Fit,
breaklinks=true}
\usepackage{cite}

\ifCLASSINFOpdf
\else
\fi
\begin{document}
	\title{Large AI Model Empowered Multimodal Semantic Communications}
	%	: More Precise Semantic Awareness and More Universal Knowledge Base}
	
	\author{Feibo Jiang, \textit{Senior Member, IEEE}, Li Dong, Yubo Peng, Kezhi Wang, \textit{Senior Member, IEEE}, Kun Yang, \textit{Fellow, IEEE}, Cunhua Pan, \textit{Senior Member, IEEE}, Xiaohu You, \textit{Fellow, IEEE}
	\thanks{
		Feibo Jiang is with Hunan Normal University, China.
		Li Dong is with the Hunan University of Technology and Business, China.
		Kezhi Wang is with Brunel University, UK.
		Yubo Peng and Kun Yang are with Nanjing University, China.
		Cunhua Pan and Xiaohu You are with Southeast University, China.
	}
	%	\thanks{The paper was submitted on \today.}\\
	}

\markboth{Submitted for Review}%
{Shell \MakeLowercase{\textit{et al.}}: Bare Demo of IEEEtran.cls for IEEE Journals}



\maketitle 

\begin{abstract}
Multimodal signals, including text, audio, image, and video, can be integrated into Semantic Communication (SC) systems to provide an immersive experience with low latency and high quality at the semantic level. However, the multimodal SC has several challenges, including data heterogeneity, semantic ambiguity, and signal distortion during transmission.
Recent advancements in large AI models, particularly in the Multimodal Language Model (MLM) and Large Language Model (LLM), offer potential solutions for addressing these issues. To this end, we propose a Large AI Model-based Multimodal SC (LAM-MSC) framework, where we first present the MLM-based Multimodal Alignment (MMA) that utilizes the MLM to enable the transformation between multimodal and unimodal data while preserving semantic consistency. Then, a personalized LLM-based Knowledge Base (LKB) is proposed, which allows users to perform personalized semantic extraction or recovery through the LLM. This effectively addresses the semantic ambiguity. Finally, we apply the Conditional Generative adversarial networks-based channel Estimation (CGE) for estimating the wireless channel state information. This approach effectively mitigates the impact of fading channels in SC.
Finally, we conduct simulations that demonstrate the superior performance of the LAM-MSC framework.
\end{abstract}

%\begin{IEEEkeywords}
	%Semantic communication; multimodality; LLM; MLM; knowledge base.
%\end{IEEEkeywords}

% \fi
\IEEEpeerreviewmaketitle

\section{Introduction}
In Weaver and Shannon's pioneering works, the communication systems can be categorized into three levels 
\cite{shannon1949mathematical}: 
\begin{enumerate}
\item Technical Level: This level emphasizes the efficiency and accuracy of the communication system, with the sender transmitting information (such as a message or signal) to the receiver. The goal is to mitigate noise or interference that could result in errors or loss of information.
	\item Semantic level: This level focuses on the meaning of the message being transmitted. The objective is to ensure that the sender and receiver understand and interpret the message in the same way. %which is crucial for effective communication.
	\item Effectiveness level: This level focuses on the impact of the communication on the receiver. The objective of the sender is to 
  accomplish its intended goal or purpose, trying to make an impact on the receiver's thoughts, behavior, or emotions.
\end{enumerate}

The rapid integration of Artificial Intelligence (AI) and wireless communications has led to the emergence of intelligent applications, such as holographic communication, and the Internet of Everything (IoE). These trends are driving the evolution of communication systems toward Semantic Communication (SC) \cite{xie2021deep}, which integrates communication with semantic information, concentrating on the ``meaning" behind transmitted bits to enable more intelligent and adaptive communication services. %Consequently, SC is capable of operating at higher levels (i.e., semantic or effectiveness levels) within Weaver's communication framework.
Typically, the SC system comprises five components, including the semantic encoder, channel encoder, channel decoder, semantic decoder, and the Knowledge Base (KB).
The KB is a structured and memory-capable knowledge network model that can provide relevant semantic knowledge descriptions for raw data. It can adopt different construction methods according to different information sources, channels, and task requirements \cite{luo2022semantic}.

Large AI models can fully leverage their immense knowledge to assist in semantic analysis and extraction, representing a cutting-edge research direction in SC.
In \cite{jiang2023semantic}, the authors aimed at the integration of Foundation Models (FMs) at the effectiveness, semantic, and physical levels, which utilized universal knowledge as a powerful tool to radically innovate system design.
G. Liu et al. \cite{10388354} introduced a comprehensive conceptual model for harmonizing AI Generated Content (AIGC) and SC, which described how AIGC and SC synergize to create content that is both meaningful and effective.
In \cite{10558819}, the authors focused on image transmission and applied the Segment Anything Model (SAM), a large vision model, to drive improvements in SC.
However, these studies did not consider the impact of personalized knowledge bases on semantic communication, and they primarily focused on the issues of semantic communication within a single modality.
%These components can be implemented by neural networks that have favorable feature extraction capabilities, and they can be trained to maximize system capacity and minimize semantic errors during transmission \cite{xie2021deep}. 

\begin{figure*}[htbp]
	\centering
	\includegraphics[width=18cm]{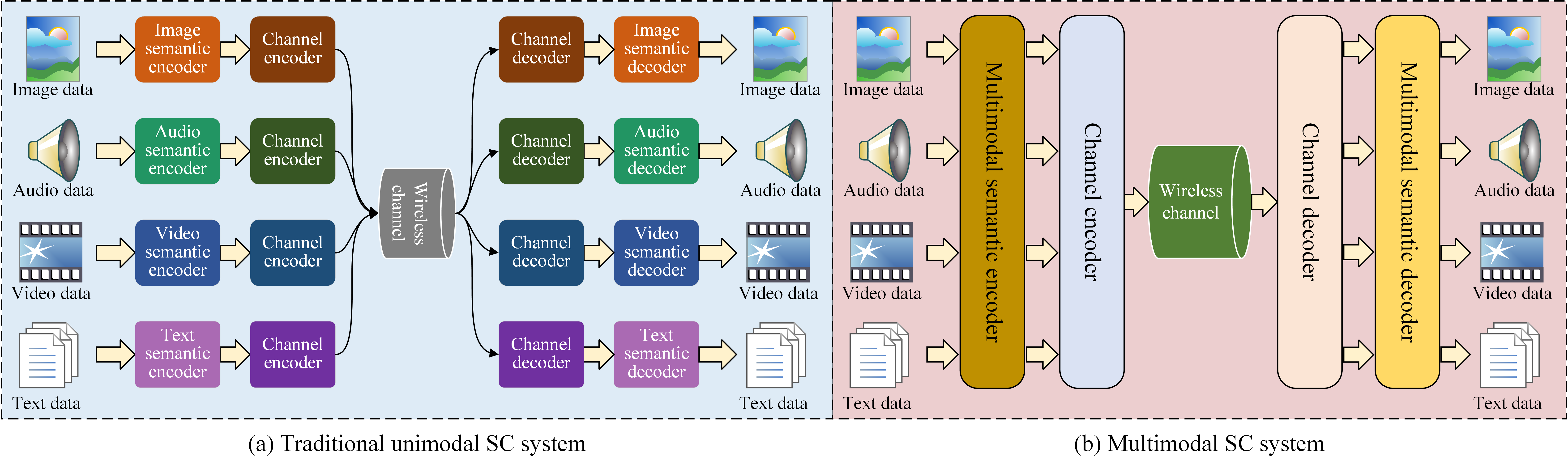}
	\caption{Traditional unimodal SC system versus multimodal SC system.}
	\label{fig:SC}
\end{figure*}
Currently, the data to be transmitted is typically multimodal for advanced applications, such as metaverse and mixed reality. As a result, the multimodal SC system is highly required to facilitate SC across multiple modes, including text, voice, images, videos, and more.
However, as illustrated in Fig. \ref{fig:SC}, (a) demonstrates that traditional SC systems are typically designed to handle only one type of unimodal data. Consequently, transmitting multimodal data requires the utilization of multiple unimodal SC systems, potentially resulting in significant overheads and inefficiencies \cite{xie2022task}. On the other hand, (b) represents a multimodal SC system capable of processing various modalities by employing a unified multimodal SC model.

\subsection{Challenges of Multimodal SC}
To better achieve multimodal SC, we summarize several challenges currently faced by multimodal SC systems:
\begin{enumerate}
\item \emph{Data heterogeneity}: 
A multimodal SC should be capable of handling the simultaneous transmission of heterogeneous data, including text, images, videos, and even specialized or rare file formats in various forms. Then, the target tasks associated with the data can be quite complex, involving machine translation, image recognition, and video analysis, among others. Additionally, semantic alignment should be considered when extracting semantic features from multimodal data, ensuring a uniform understanding across different multimodal data.

\item \emph{Semantic ambiguity}:
On one hand, multimodal SC systems may encounter issues such as semantic errors or misunderstandings when transmitting multimodal data from one modality to another, resulting in the semantic ambiguity. On the other hand, each party in communication has distinct knowledge backgrounds and may focus on different semantic information. This may cause an inconsistent understanding of the semantic information between different parties, contributing to semantic ambiguity.
	
\item \emph{Signal distortion}:
The signal transmission may be impacted by fading/noisy channels over time, influenced by factors like environmental conditions.
%, distance, and interference. 
This fluctuation adds a layer of complexity to the accurate and meaningful exchange of information between senders and receivers. In other words, wireless channels may incur transmitting signal disturbances \cite{xie2021deep} causing the loss of critical information or the alteration of intended semantics, further complicating the process of re-establishing personalized semantics.
\end{enumerate} 

\subsection{Advantages of Large AI Model in Multimodal SC}
Recent advancements in Deep Learning (DL) have enabled the development of large AI models for multimodal data and Natural Language Processing (NLP), resulting in models with enhanced capabilities in these domains, such as Multimodal Language Model (MLM), e.g., Composable Diffusion (CoDi) \cite{tang2023any} and Gemini, and Large Language Model (LLM), e.g., GPT-4 \cite{achiam2023gpt}. These large AI models have the following advantages for SC:
\begin{itemize}
	\item {\it{Accurate Semantic Extraction}}: With billions of parameters, large AI models can learn intricate representations, providing high-quality semantic extraction of input data.
	
	\item {\it{Rich Prior/Background Knowledge}}:
	Pre-trained on vast datasets like ImageNet, Audioset, and Wikipedia, large AI models gain extensive domain knowledge, exhibiting excellent world model capabilities.
	
	\item {\it{Robust Semantic Interpretation}}:
	With their robust generation capabilities, large AI models can effectively interpret diverse semantic information, even when faced with semantic noise.
\end{itemize}

%Furthermore, the continuous enhancement of computational power and the growth in data volume have allowed these sophisticated AI models to be successfully implemented across various industries, providing potential solutions for achieving multimodal SC.

\subsection{Our Contributions}
We propose a Large AI Models-based Multimodal SC (LAM-MSC) framework to address the above-mentioned challenges. Our contributions can be summarized as follows:
\begin{enumerate}
\item \emph{Unified Semantic Representation}: 
We introduce an MLM-based Multimodal Alignment (MMA) by employing CoDi for modality transformation. MMA facilitates the synchronized generation of interwoven modalities by constructing a shared multimodal space. Since the same semantics are represented in different forms in different modal data, we unify the multimodal data into the text modality because it can represent the semantics accurately using the minimum data volume. This approach aims to enhance the efficiency of multimodal SC systems while ensuring semantic consistency.
	
	\item \emph{Personalized Semantic Understanding}: We propose an LLM-based Knowledge Base (LKB) utilizing the GPT-4 model to understand personal information. Specifically, we design a personalized prompt base, which includes various personalized information such as individual profiles. Prompt learning is employed to finetune the global GPT-4 model using the personalized prompt base, thereby creating a personalized local KB. The personalized KB can extract and analyze more relevant semantic information and eliminate semantic ambiguities. 
	
	\item \emph{Generative Channel Estimation}: We train and employ Conditional Generative Adversarial Networks-based Channel Estimation (CGE) to estimate channel gains of fading channels, utilizing pilot sequence as the conditional information fed into the network. Considering the characteristics of channel gains, we design a dedicated generator network based on convolution and deconvolution structures. We also employ a leakyReLU activation function to capture the nonlinear properties and generate high-quality channel gains.
	
\end{enumerate}

The rest part is structured as follows: First, we introduce the CoDi for multimodal data and GPT-4 for personalized KB. 
Next, we present the LAM-MSC framework and its key components, including MMA, LKB, and CGE methods. Subsequently, we provide simulations to evaluate the performance of the LAM-MSC framework. Finally, we conclude the paper.

\section{Preliminaries}
\subsection{CoDi for Multimodal Data}
CoDi is an innovative MLM introduced by Microsoft, capable of generating output modalities (text, image, video, audio) from any combination of input modalities. The key components of CoDi include \cite{tang2023any}:

\subsubsection{Latent Diffusion Process}
Unlike traditional diffusion models that operate directly in the data space, latent diffusion begins by encoding data into a compact and latent representation. This latent representation is then guided by the learned diffusion model to reconstruct high-quality output in the latent space before decoding it back into the data space.

\subsubsection{Unimodal Module Design}
Different modalities or conditions of the generation task are encapsulated in separate modules. These modules can encapsulate a variety of information or constraints, such as textual descriptions, image features, or specific attributes that the generated content should adhere to.

\subsubsection{Composable Multimodal Condition}
During the generation process, CoDi adeptly combines modalities or conditions from its various modules to guide the denoising process in the latent space. This composition enables the flexible integration of multiple, potentially diverse, modalities or conditions into a single generative process.

\subsubsection{Reverse Multimodal Generation}
Leveraging the latent diffusion denoising, CoDi generates content through a process that incrementally removes noise while integrating the composited conditions, optimizing the reconstructed latent representation to ensure the model generates content that aligns with semantic representations of different modalities or conditions.

\subsection{GPT-4 for Personalized KB}
\subsubsection{GPT-4-Based Global KB}
GPT-4, introduced by OpenAI in 2023 \cite{achiam2023gpt}, is among the most advanced LLMs, succeeding GPT-3 and GPT-3.5 as the latest evolution in the GPT series. This model adopts the transformer architecture and boasts approximately 100 billion parameters. Trained on vast text corpora containing trillions of words, GPT-4 excels at learning intricate language representations. The model's capabilities in multi-modal knowledge synthesis, semantic summarization, continuous learning, and scalability make it highly suitable for automatically populating and expanding KBs from unstructured data. As a result, GPT-4 is utilized as the global KB. While GPT-4-based global KBs are built on general textual data, fine-tuning enables them to adapt to more specialized domains, such as medicine, finance, or communication. 

\subsubsection{Fine-Tuning-Based Personalized KB}
Large AI models can be updated with few samples, allowing adaptation to specific tasks such as personalized applications. There are four primary fine-tuning methods to transform the GPT-4-based global KB into a personalized KB for individuals \cite{phang2023hypertuning}:
\begin{itemize}
	\item \emph{Adapter Tuning} trains a few parameters in small networks called adapter modules, inserted after each layer in the original LLM. By fixing pre-trained model parameters and training only adapter module parameters, computational costs are reduced while preserving pre-training knowledge.
	
	\item \emph{Prefix Tuning} is a parameter-efficient method that trains a small set of parameters called the ``prefix" to modify the input for the pre-trained model. The prefix optimizes task-specific input, requiring less computational resources than full model fine-tuning.
	
	\item \emph{Prompt Tuning} allows users to guide the behavior of LLMs and align their responses by prompt for specific requirements or objectives. By carefully designing and refining prompts, it is possible to improve the quality, relevance, and accuracy of the generated outputs.
	
	\item \emph{Low-Rank Adaptation (LoRA)} aims for transparent and interpretable fine-tuning by adding a low-rank matrix to each pre-trained model layer, and fine-tuning it for target tasks while keeping the original pre-trained weights fixed. 
\end{itemize}

\subsection{CGAN for Channel Estimation}
Channel estimation is a crucial task in wireless communication systems, involving the prediction of vital channel characteristics such as channel gains %, phase offsets, and frequency offsets
based on received data \cite{9246559}. Accurate channel estimation is essential for the receiver to effectively reconstruct the transmitted signal, leading to improved communication efficiency and quality.

It is worth noting that the pilot sequence, received signal and channel gains can be treated as dual-channel images, where each image represents the real and imaginary components of a complex matrix. Hence, the task of channel estimation can be reframed as an image-to-image translation problem \cite{9246559}. Consequently, Conditional Generative Adversarial Networks (CGAN) can be leveraged for channel estimation. In this approach, the generator is trained to learn the mapping relationship between the received signal, pilot sequence, and channel gains. Simultaneously, the discriminator plays a role in distinguishing the generated channel gains, thereby aiding in the improvement of the generator's performance.

\section{Implementation of Multimodal SC}
\begin{figure*}[htbp]
	\centering
	\includegraphics[width=18cm]{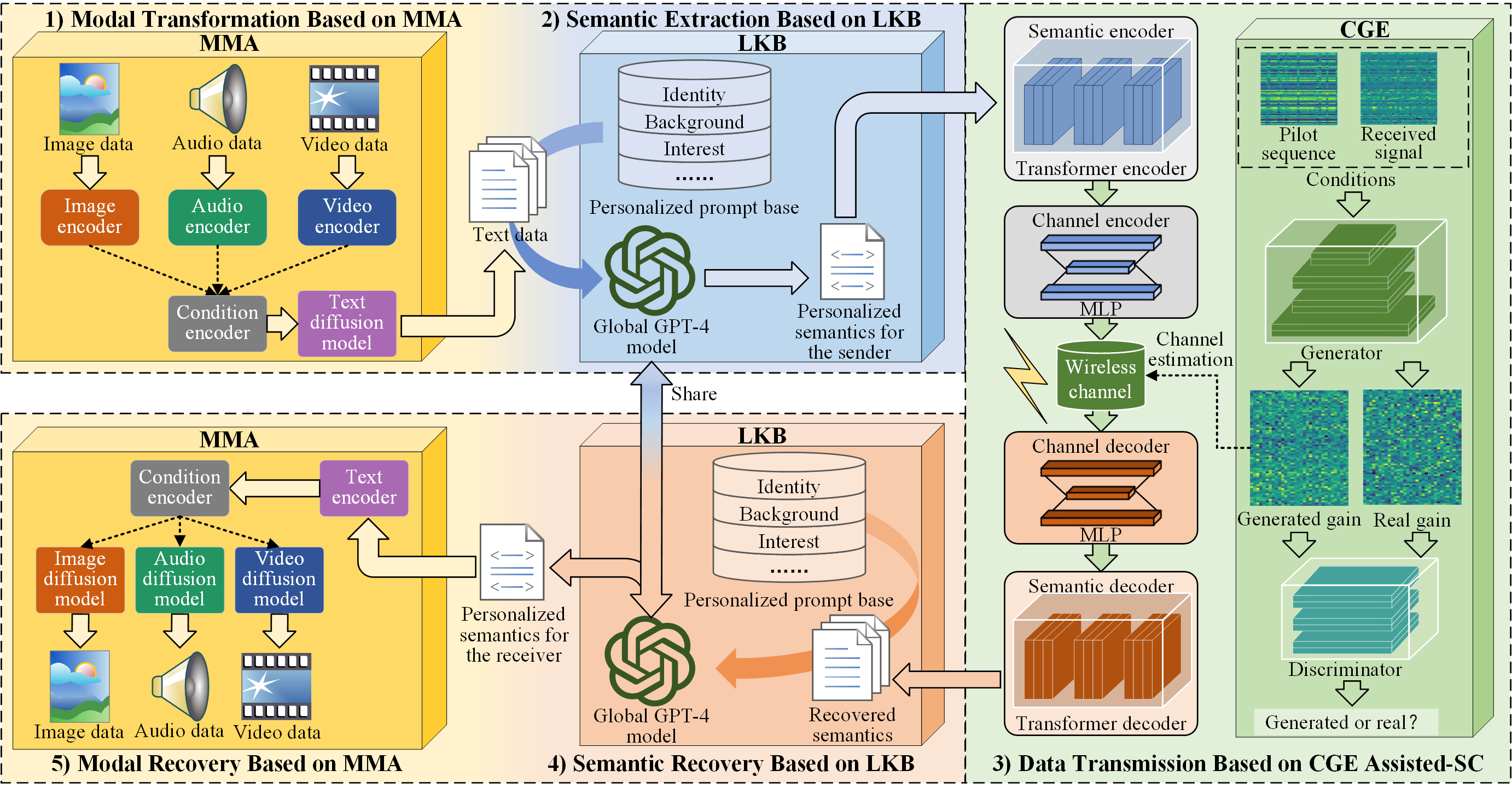}
\caption{The workflow of the proposed LAM-MSC framework.} 
	\label{fig:LAM-MSC}
\end{figure*}
\begin{figure*}[htbp]
	\centering
	\includegraphics[width=18cm]{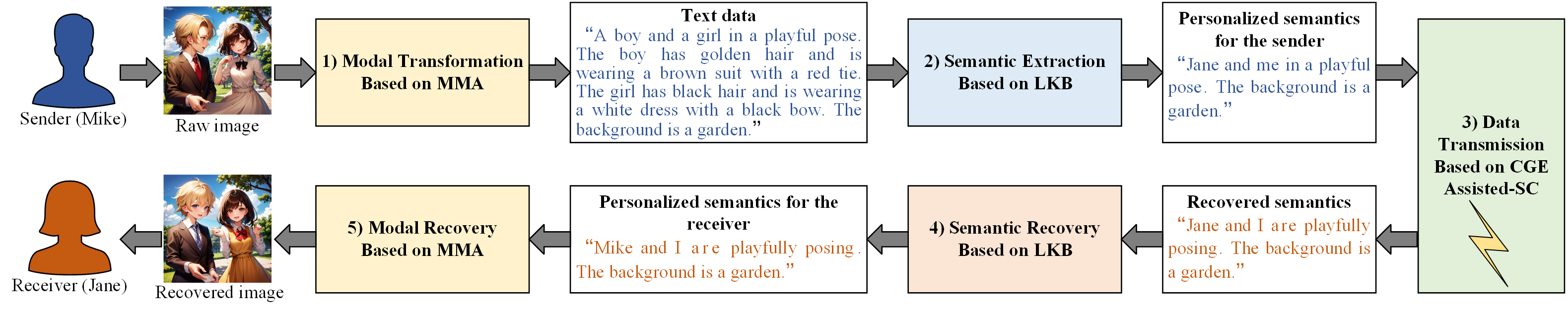}
	\caption{A dataflow example of the proposed LAM-MSC framework: Sender Mike dispatches an image to receiver Jane with the intention of conveying the semantic content of the image as ``Mike and Jane are playing in a garden."}
	\label{fig:eg}
\end{figure*}

%We propose the LAM-MSC framework, leveraging the power of large AI models to solve the previously mentioned challenges (i.e., data heterogeneity, semantic ambiguity, and signal fading). 
The key to the LAM-MSC framework is that we introduce the CoDi model to facilitate the transformation of heterogeneous multimodal data into a singular unimodal format. We choose text data as the unimodal format due to its various benefits, including human readability, high information density, limited redundancy, and lower storage demands compared to video or audio formats.
Information density represents the amount of semantic information contained per unit of data, which is the ratio of the amount of semantic information to the amount of raw data in an unimodal space \cite{he2022masked}. 
Moreover, using text data as the unimodal format enables us to apply GPT-4 as the KB, enhancing the accuracy of semantic extraction and the interpretability of data recovery.

\subsection{LAM-MSC Framework}
To implement the multimodal SC, we adopt the LAM-MSC framework, which integrates large AI models as a solution. In this framework, MMA utilizes the CoDi model to facilitate the conversion between multimodal data and textual data. Then, LKB leverages a personalized prompt base and GPT-4 to enhance the understanding and disambiguation of personal information. Additionally, CGE is employed to estimate channel gains in wireless channels. As shown in Fig. \ref{fig:LAM-MSC}, the workflow of the LAM-MSC framework is summarized as follows:

\subsubsection{Modal Transformation Based on MMA}
For the input multimodal data, which may include image, audio, and video data, MMA is utilized to convert these data into text data while maintaining semantic alignment. The corresponding text data can effectively capture the original modal data's content. 
For example, as illustrated in Fig. \ref{fig:eg}, the raw data consists of a photograph featuring the sender (assumed to be Mike) and the receiver (assumed to be Jane) playing in a garden. The raw image is then converted into a text description: ``A boy and a girl in a playful pose. The boy has golden hair and is wearing a brown suit with a red tie. The girl has black hair and is wearing a white dress with a black bow. The background is a garden".
Thus, by applying MMA, we manage to transform multimodal data into unimodal data while ensuring semantic alignment.

\subsubsection{Semantic Extraction Based on LKB}
For the text data obtained through modal transformation, senders typically aim to transmit only the key information that expresses their intended message or the parts they find most important while omitting redundant information they deem irrelevant for the receiver. This personalized key information can be referred to as semantics. Hence, LKB is used to personalize the text and thus obtain personalized semantics. 
As illustrated in Fig. \ref{fig:eg}, the raw text initially lacks personalized information. However, through the integration of the sender's intention, user information, and interests, the LKB extracts personalized semantics ``Jane and me in a playful pose. The background is a garden." This description encompasses the identities of the sender and receiver and indicates that the sender's focus primarily involves the ``two people" and the ``place" depicted in the image, rather than other details like attire or clothes.

\subsubsection{Data Transmission Based on CGE Assisted-SC}
SC starts with a semantic encoder that extracts meaningful elements or attributes from raw data, aiming to transmit this semantic information as accurately as possible to the receiver. Then, the channel encoder modulates the semantically encoded data into complex-valued input symbols suitable for wireless transmission.
To mitigate the effects of the fading channel, the CGE is employed to acquire the channel gains, which can reduce the complexity involved in the channel decoder's recovery of transmitted signals.
Next, the channel decoder is utilized to perform signal demodulation while overcoming the additive noise.
Finally, the semantic decoder performs semantic decoding to retrieve recovered semantics (e.g., ``Jane and I are playfully posing. The background is a garden"). %Although the physical channel impairments cause slight differences between recovered semantics and original content, overall meaning consistency is maintained.}

\subsubsection{Semantic Recovery Based on LKB}
The receiver may not understand the recovered semantics directly since the personalization of received messages is specific to the sender rather than the receiver, which can lead to semantic ambiguous issues. 
Hence, similarly, the LKB is adopted to change the decoded semantics into the personalized semantics for the receiver according to the personalized prompt base of the receiver.
As shown in Fig. \ref{fig:eg}, the LKB adapts the recovered semantics based on the receiver's user information, such as their identity. As a result, the recovered semantics are customized and transformed into personalized semantics for the receiver, Jane, resulting in the text ``Mike and I are playfully posing. The background is a garden".

\subsubsection{Modal Recovery Based on MMA}
Similar to modal transformation, MMA is utilized to achieve modal recovery, meaning it converts text data back into the original modal data. However, it is important to note that we only evaluate the consistency between the recovered and original modal data in terms of semantics rather than bits at the data level. 
As illustrated in Fig. \ref{fig:eg}, the recovered image displays the scene as ``Mike and Jane are playing in a garden." This is a result of the sender's primary intention, which focuses on the semantic aspect of the characters and background, rather than providing specific details about clothing or other elements.

\subsection{MMA}
In the proposed LAM-MSC framework, MMA performs the multimodal transformation. As shown in Fig. \ref{fig:LAM-MSC}, the workflow of MMA can be summarized below:
 
\subsubsection{Modal Transformation}
On the sender side, the MMA transforms multimodal data, including image, audio, and video data, into unimodal textual data. Specifically, each type of multimodal data is first encoded by its respective encoder. 
Then, the encoding results of the multimodal data are fed into the condition encoder, which processes them according to the target modality being transitioned to, in this case, the text modality. Finally, the processed results from the condition encoder are input into the text diffusion model to generate corresponding textual data that maintains semantic consistency with the original multimodal data.

\subsubsection{Modal Recovery}
On the receiver side, the MMA facilitates the transformation of personalized semantics (e.g., textual data) back into the original multimodal data. Specifically, 
the personalized semantics are first fed into the text encoder to obtain the text features. Then, the text features are input into the conditional encoder, which processes the data based on the target modality being recovered, such as image, audio, and video data. 
Finally, the processed result from the conditional encoder is input into the diffusion model of the target modality, which encompasses image, audio, and video diffusion models. This generates corresponding modality data that ensures semantic consistency with the input personalized semantics.

\subsection{LKB}
LKB primarily consists of two components: The global GPT-4 model and the personalized prompt base. The descriptions of these components are summarized below:
\subsubsection{Global GPT-4 KB}
The GPT-4 model boasts outstanding capabilities in NLP, allowing it to perform precise semantic extraction and restoration from textual data according to specific requirements. 
With numerous parameters and multi-head attention mechanisms, GPT-4 excels at accurate knowledge representation, allowing it to comprehend semantics and knowledge structures with precision. Additionally, GPT-4 has been pre-trained using extensive datasets, which makes it store rich prior/background knowledge and achieve strong generalization abilities across different domains.
Hence, the GPT-4 model is used as the shared global KB for all users, serving as a ``global" model consistently utilized across a diverse array of applications.

\subsubsection{Personalized Prompt Base}
As discussed in Section II-B, there are four primary methods for achieving personalization in GPT-4 models. However, methods such as adapter tuning, prefix tuning, and LoRA involve adjusting the GPT-4 model's structure. These modifications necessitate users to possess specific professional knowledge and require their devices to be equipped with substantial resource support. Clearly, this is an unrealistic demand for the majority of normal users.

Therefore, we utilize prompt tuning in combination with a personalized prompt base to fine-tune the GPT-4 model. The personalized prompt base includes character profiles, such as names, ages, identities, genders, interests, and other information, which can be easily organized in a tabular format (as illustrated in Fig. \ref{fig:LAM-MSC}).
As a result, users only need to input this prompt base along with the text data into the global GPT-4 model, after which the personalized semantics are generated.

\subsection{CGE} 
As illustrated in Fig. \ref{fig:LAM-MSC}, we utilize the CGE to estimate wireless channel gains. This information greatly enhances the accuracy of semantic transmission in wireless channels. 
Specifically, we propose using CGAN to estimate channel gains according to received signals and pilot sequences. The CGAN consists of a generator and a discriminator during the training phase. 
The generator includes three downsampling blocks with convolutional layers, two upsampling blocks with deconvolutional layers, and an output layer. The convolutional and deconvolutional layers are applied to capture the local features of the channel gains. We also introduce a novel LeakyReLU activation function to model the nonlinear characteristics of the channel gains.
The discriminator consists of four convolutional layers with ReLU activation functions.

Upon completion of the adversarial training, the trained generator can be utilized to estimate wireless channel information, i.e., gains from the conditional inputs (i.e., the received signals and pilot sequences), mitigating the influence of fading channels in the SC system.

\section{Simulation Results}
\subsection{Problem Formulation}
We focus on an end-to-end data communication scenario that encompasses the transmission of various data types, including images, audio, and videos.
These multimodal data are transformed into unimodal data (i.e., textual data) by MMA. 
Moreover, we employ BERT and cosine similarity to evaluate the performance of the multimodal SC system \cite{zhang2019bertscore}.
BERT is a pre-trained foundational model proposed by Google for high-quality semantic encoding of textual data.
Cosine similarity is a mathematical method used to measure the similarity between two semantic vectors produced by BERT. Its range is from -1 to 1, where -1 means complete opposites, and 1 means the same \cite{zhang2019bertscore}.
Then, a predetermined cosine similarity threshold is used to assess the accuracy of SC.

\subsection{Simulation Settings}
First, we present the evaluation datasets for the multimodal SC as follows:
\begin{itemize}
	\item VOC2012 (image dataset): This dataset comprises 17,125 RGB images across 20 categories.
	
	\item LibriSpeech (audio dataset): This corpus contains approximately 1,000 hours of 16 kHz English speech readings.
	
	\item UCF101 (video dataset): This action recognition dataset consists of realistic action videos from YouTube, spanning 101 action categories.
\end{itemize}

Second, the SC model is designed for textual modal data. Thus, we apply the transformer as the network architecture. The channel model, which encompasses channel encoding and decoding along with wireless channel configuration, adopts settings similar to those presented in \cite{xie2021deep}. 

Finally, the threshold for cosine similarity is set at $0.6$. This indicates that the transmitted semantics are considered to be accurate only when the cosine similarity between the textual features exceeds $0.6$. The transmission accuracy is defined as the ratio of semantically correct samples to the total number of transmitted samples (i.e., the sum of the texts converted by the three modalities).

\subsection{Evaluation Results}
\begin{figure}[htbp]
	\centering
	\includegraphics[width=8cm]{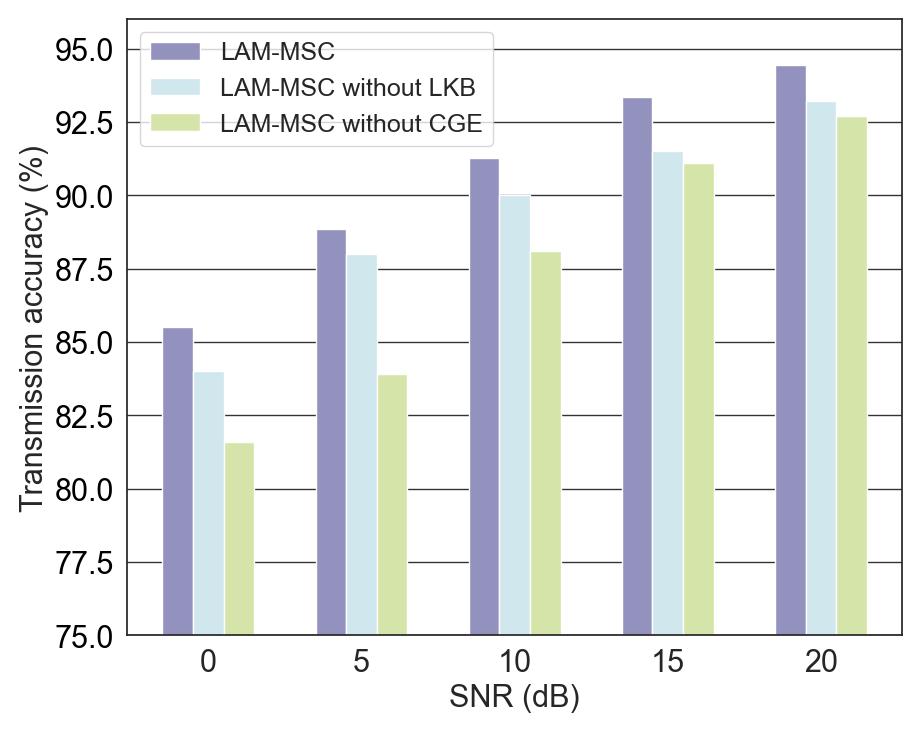}
	\caption{Transmission accuracy of multimodal SC under different SNRs.}
	\label{fig:exp1}
\end{figure}

The results of ablation experiments are illustrated in Fig. \ref{fig:exp1}, where we observe that the transmission accuracy of multimodal SC increases as the SNR improves. 
One can see that the personalized prompt can improve the accuracy of semantic transmission when comparing LAM-MSC and LAM-MSC without LKB.
Furthermore, one can also see that the performance of LAM-MSC without CGE is the worst, indicating the importance of having CGE in the proposed SC system.  

\begin{figure}[htbp]
	\centering
	\includegraphics[width=8.5cm]{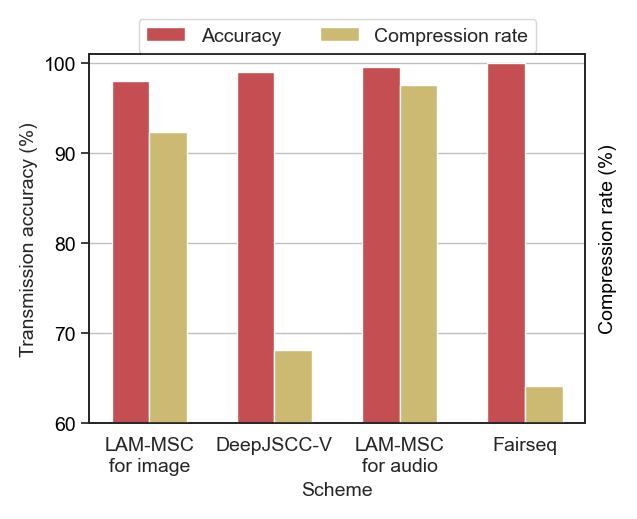}
	\caption{Comparison results of different schemes.}
	\label{fig:exp2}
\end{figure}

Fig. \ref{fig:exp2} depicts the results of comparison experiments, where we evaluate DeepJSCC-V \cite{10015684} for image transmission and Fairseq \cite{ott2019fairseq} for audio transmission as contenders. 
Additionally, the compression rate in Fig. \ref{fig:exp2} is defined as the ratio between compressed data and original data. This means that less transmitted data indicates a higher compression rate.
Since DeepJSCC-V and Fairseq are specifically designed for their respective single modalities, they slightly surpass LAM-MSC in terms of transmission accuracy. 
However, since the LAM-MSC can convert the image and audio to textual data and thus the required transmitted semantic information is reduced, the LAM-MSC exhibits significant advantages in terms of the compression ratio. 
Moreover, DeepJSCC-V and Fairseq may only process unimodal data, whereas the proposed LAM-MSC is capable of effectively handling multimodal information. 


\section{Open Issues}

\subsubsection{Unified Representation}
Although the conversion of multimodal data into unimodal data is considered in this paper, developing a comprehensive and universal semantic representation for more modalities still proves challenging. An effective method to consistently represent multimodal data would enhance interoperability and comprehension across various modalities.

\subsubsection{Semantics Compression}
Multimodal data could be extensive, necessitating the implementation of efficient compression techniques for transmission. The preservation of semantic information during the data compression process represents an open issue, as conventional methods may contribute to the loss of vital context.

\subsubsection{Noise Robustness}
Multimodal data sources may contain noise, which could diminish the performance of SC systems. The development of algorithms and methods for enhancing robustness and maintaining SC quality among varying environments remains important.

\subsubsection{Adaptability and Scalability}
With the rapid growth of data volume and diverse demands, the next generation of SC might require flexible and scalable approaches that can effectively manage and process extensive and multimodal data. 

\section{Conclusion}
In this paper, we first introduced the challenges faced by multimodal SC. Then, we presented a LAM-MSC framework that incorporates MMA, enabling transformations between multimodal and unimodal data while preserving semantic consistency. Next, a personalized LKB was proposed in LAM-MSC, allowing users to undertake individualized semantic extraction or recovery, effectively tackling semantic ambiguous issues in transmitted data. Additionally, we applied CGE to estimate the wireless channel gains which can reduce the impact of fading channels in SC. Finally, simulations demonstrated the superior performance of the LAM-MSC framework in processing multimodal SC systems.

\section*{Acknowledgments}
This work was supported in part by the
National Natural Science Foundation of China under Grant 41904127 and 62132004, in part by the Hunan Provincial Natural Science Foundation of China under Grant 2024JJ5270, in part by the Open Project of Xiangjiang Laboratory under Grant 22XJ03011, and in part by the Scientific Research Fund of Hunan Provincial Education Department under Grant 22B0663.

\bibliographystyle{ieeetran}
\bibliography{bare_jrnl_bobo}

\begin{thebibliography}{10}
\providecommand{\url}[1]{#1}
\csname url@samestyle\endcsname
\providecommand{\newblock}{\relax}
\providecommand{\bibinfo}[2]{#2}
\providecommand{\BIBentrySTDinterwordspacing}{\spaceskip=0pt\relax}
\providecommand{\BIBentryALTinterwordstretchfactor}{4}
\providecommand{\BIBentryALTinterwordspacing}{\spaceskip=\fontdimen2\font plus
\BIBentryALTinterwordstretchfactor\fontdimen3\font minus
  \fontdimen4\font\relax}
\providecommand{\BIBforeignlanguage}[2]{{%
\expandafter\ifx\csname l@#1\endcsname\relax
\typeout{** WARNING: IEEEtran.bst: No hyphenation pattern has been}%
\typeout{** loaded for the language `#1'. Using the pattern for}%
\typeout{** the default language instead.}%
\else
\language=\csname l@#1\endcsname
\fi
#2}}
\providecommand{\BIBdecl}{\relax}
\BIBdecl

\bibitem{shannon1949mathematical}
C.~E. Shannon and W.~Weaver, \emph{The Mathematical Theory of
  Information}.\hskip 1em plus 0.5em minus 0.4em\relax University of illinois
  Press, 1949.

\bibitem{xie2021deep}
H.~Xie, Z.~Qin, G.~Y. Li, and B.-H. Juang, ``Deep learning enabled semantic
  communication systems,'' \emph{IEEE Transactions on Signal Processing},
  vol.~69, pp. 2663--2675, 2021.

\bibitem{luo2022semantic}
X.~Luo, H.-H. Chen, and Q.~Guo, ``Semantic communications: Overview, open
  issues, and future research directions,'' \emph{IEEE Wireless
  Communications}, vol.~29, no.~1, pp. 210--219, 2022.

\bibitem{jiang2023semantic}
P.~Jiang, C.-K. Wen, X.~Yi, X.~Li, S.~Jin, and J.~Zhang, ``Semantic
  communications using foundation models: Design approaches and open issues,''
  \emph{arXiv preprint arXiv:2309.13315}, 2023.

\bibitem{10388354}
G.~Liu, H.~Du, D.~Niyato, J.~Kang, Z.~Xiong, D.~I. Kim, and X.~Shen, ``Semantic
  communications for artificial intelligence generated content ({AIGC}) toward
  effective content creation,'' \emph{IEEE Network}, pp. 1--1, 2024.

\bibitem{10558819}
F.~Jiang, Y.~Peng, L.~Dong, K.~Wang, K.~Yang, C.~Pan, and X.~You, ``Large {AI}
  model-based semantic communications,'' \emph{IEEE Wireless Communications},
  vol.~31, no.~3, pp. 68--75, 2024.

\bibitem{xie2022task}
H.~Xie, Z.~Qin, X.~Tao, and K.~B. Letaief, ``Task-oriented multi-user semantic
  communications,'' \emph{IEEE Journal on Selected Areas in Communications},
  vol.~40, no.~9, pp. 2584--2597, 2022.

\bibitem{tang2023any}
Z.~Tang, Z.~Yang, C.~Zhu, M.~Zeng, and M.~Bansal, ``Any-to-any generation via
  composable diffusion,'' \emph{arXiv preprint arXiv:2305.11846}, 2023.

\bibitem{achiam2023gpt}
J.~Achiam, S.~Adler, S.~Agarwal, L.~Ahmad, I.~Akkaya, F.~L. Aleman, D.~Almeida,
  J.~Altenschmidt, S.~Altman, S.~Anadkat \emph{et~al.}, ``Gpt-4 technical
  report,'' \emph{arXiv preprint arXiv:2303.08774}, 2023.

\bibitem{phang2023hypertuning}
J.~Phang, Y.~Mao, P.~He, and W.~Chen, ``Hypertuning: Toward adapting large
  language models without back-propagation,'' in \emph{International Conference
  on Machine Learning}.\hskip 1em plus 0.5em minus 0.4em\relax PMLR, 2023, pp.
  27\,854--27\,875.

\bibitem{9246559}
Y.~Dong, H.~Wang, and Y.-D. Yao, ``Channel estimation for one-bit multiuser
  massive {MIMO} using conditional gan,'' \emph{IEEE Communications Letters},
  vol.~25, no.~3, pp. 854--858, 2021.

\bibitem{he2022masked}
K.~He, X.~Chen, S.~Xie, Y.~Li, P.~Doll{\'a}r, and R.~Girshick, ``Masked
  autoencoders are scalable vision learners,'' in \emph{Proceedings of the
  IEEE/CVF conference on computer vision and pattern recognition}, 2022, pp.
  16\,000--16\,009.

\bibitem{zhang2019bertscore}
T.~Zhang, V.~Kishore, F.~Wu, K.~Q. Weinberger, and Y.~Artzi, ``Bertscore:
  Evaluating text generation with bert,'' \emph{arXiv preprint
  arXiv:1904.09675}, 2019.

\bibitem{10015684}
W.~Zhang, H.~Zhang, H.~Ma, H.~Shao, N.~Wang, and V.~C.~M. Leung, ``Predictive
  and adaptive deep coding for wireless image transmission in semantic
  communication,'' \emph{IEEE Transactions on Wireless Communications},
  vol.~22, no.~8, pp. 5486--5501, 2023.

\bibitem{ott2019fairseq}
M.~Ott, S.~Edunov, A.~Baevski, A.~Fan, S.~Gross, N.~Ng, D.~Grangier, and
  M.~Auli, ``fairseq: A fast, extensible toolkit for sequence modeling,''
  \emph{arXiv preprint arXiv:1904.01038}, 2019.

\end{thebibliography}
\vspace{-0.3cm}
\section*{Biographies}

\textbf{Feibo Jiang} (jiangfb@hunnu.edu.cn) received Ph.D. degree from the Central South University, China. He is currently an Associate Professor at Hunan Normal University, China.

\textbf{Li Dong} (Dlj2017@hunnu.edu.cn) received Ph.D. degree from the Central South University, China. She is currently an Associate Professor at Hunan University of Technology and Business, China.

\textbf{Yubo Peng} (pengyubo@hunnu.edu.cn) is currently pursuing a Ph.D. degree at Nanjing University, China. 

\textbf{Kezhi Wang} (Kezhi.Wang@brunel.ac.uk) received Ph.D. degree from University of Warwick, U.K. in 2015. Currently he is a Senior Lecturer with the Department of Computer Science, Brunel University London, U.K.

\textbf{Kun Yang} (kyang@ieee.org) received his PhD from the Department of Electronic \& Electrical Engineering of University College London (UCL), U.K. He is currently a Chair Professor in the School of Intelligent Software and Engineering, Nanjing University, China.

\textbf{Cunhua Pan} (cpan@seu.edu.cn) received Ph.D. degrees from Southeast University, China, in 2015. 
He is a full professor in Southeast University, China. 

\textbf{Xiaohu You} (xhyu@seu.edu.cn) received the M.S. and Ph.D. degrees in electrical engineering from Southeast University, Nanjing, China, in 1985 and 1988, respectively. 
He was a recipient of the National 1st Class Invention Prize in 2011. 
He is an Academician of the Chinese Academy of Sciences. 
\newpage
\end{document}